\pdfoutput=1

\documentclass[11pt]{article} 

\usepackage[]{emnlp2021}

\usepackage{times}
\usepackage{latexsym}

\usepackage[T1]{fontenc}

\usepackage[utf8]{inputenc}

\usepackage{times}
\usepackage{graphicx}
\usepackage{wrapfig}
\usepackage{dsfont}
\usepackage{latexsym}
\usepackage[linesnumbered,algoruled,boxed,noend]{algorithm2e}
\usepackage{amssymb}
\usepackage{amsmath}
\usepackage{url} 
\SetKwInOut{Parameter}{parameter}
\usepackage{enumitem}
\usepackage{mathtools}
\usepackage{cleveref}
\usepackage{shortcuts}
\usepackage{multirow}
\usepackage{hhline}
\usepackage[export]{adjustbox}
\usepackage{booktabs,array}
\usepackage{amsmath, bm}
\usepackage{color, colortbl}
\usepackage{booktabs,makecell,tabularx}

\author{ 
Dong-Ho Lee\textsuperscript{1,2},~
Zhiqiang Hu\textsuperscript{3},~
Roy Ka-Wei Lee\textsuperscript{3},~

\\
\textsuperscript{1}Department of Computer Science, University of Southern California \\
\textsuperscript{2}Upstage AI Research, Republic of Korea \\
\textsuperscript{3}Singapore University of Technology and Design

\\
{\texttt{dongho.lee@usc.edu, zhiqiang\_hu@mymail.sutd.edu.sg,}} \\
{\texttt{roy\_lee@sutd.edu.sg}}
} 

\begin{document}
\title{Improving Text Auto-Completion with Next Phrase Prediction}
\maketitle
\begin{abstract}
Language models such as GPT-2 have performed well on constructing syntactically sound sentences for text auto-completion task. However, such models often require considerable training effort to adapt to specific writing domains (e.g., medical). In this paper, we propose an intermediate training strategy to enhance pre-trained language models' performance in the text auto-completion task and fastly adapt them to specific domains. Our strategy includes a novel self-supervised training objective called \textit{Next Phrase Prediction (NPP)}, which encourages a language model to complete the partial query with enriched phrases and eventually improve the model's text auto-completion performance. Preliminary experiments have shown that our approach is able to outperform the baselines in auto-completion for email and academic-writing domains.
\end{abstract}
\section{Introduction}
Natural language interface (NLI) applications such as  Personal assistants (e.g., Amazon Alexa, Apple Siri, Google Assistant, and Microsoft Cortana) and search engines (e.g., Google) have become an integral part of our everyday life. 
Among the many features in NLI applications, \textit{text auto-completion}, which aims to suggest words, phrases, and sentences that complete the user's textual input, is a common, but key feature. \textit{Smart reply}~\cite{kannan2016smart} and \textit{Smart compose}~\cite{chen2019gmail} are two recent works that provide contextual assistance to aid users in completing everyday text such as emails, search engine inputs, etc.

While recent advances in deep neural models have shown impressive performance on the text auto-completion task, these models generally require a large amount of everyday text and huge amount of computing power for training to generate adequate suggestions~\cite{chen2019gmail}.
The challenge is compounded when we perform auto-completion in specific domains such as academic writing, which requires a large training corpus for specific expertise. ~\tabref{tab:perplexity} illustrates the difficulty in domain-specific auto-completion with the same amount of supervisions. 

\begin{table}[!tb]
\begin{minipage}{0.5\textwidth}
\centering
\small

\resizebox{0.8\textwidth}{!}{
	\begin{tabular}{c|cc}
		\toprule
		  \textbf{Test Perplexity} & Email & Academic Writing \\
		\midrule
		  Bi-LSTM & 1.88 $\pm$ 0.05 & 3.17 $\pm$ 0.03 \\ 
		\bottomrule
	\end{tabular}
	}
	\end{minipage}
    \vspace{0.1cm}
	\caption{\textbf{Perplexity Comparison of Bi-LSTM.} Train Bi-LSTM for the language modeling with the same amounts (100K) of training instances for each domain. Perplexity of Academic writing domain is almost double of emails.}
	\label{tab:perplexity}
\end{table}

\begin{figure}[t]
\centering
	\includegraphics[width=0.9\linewidth]{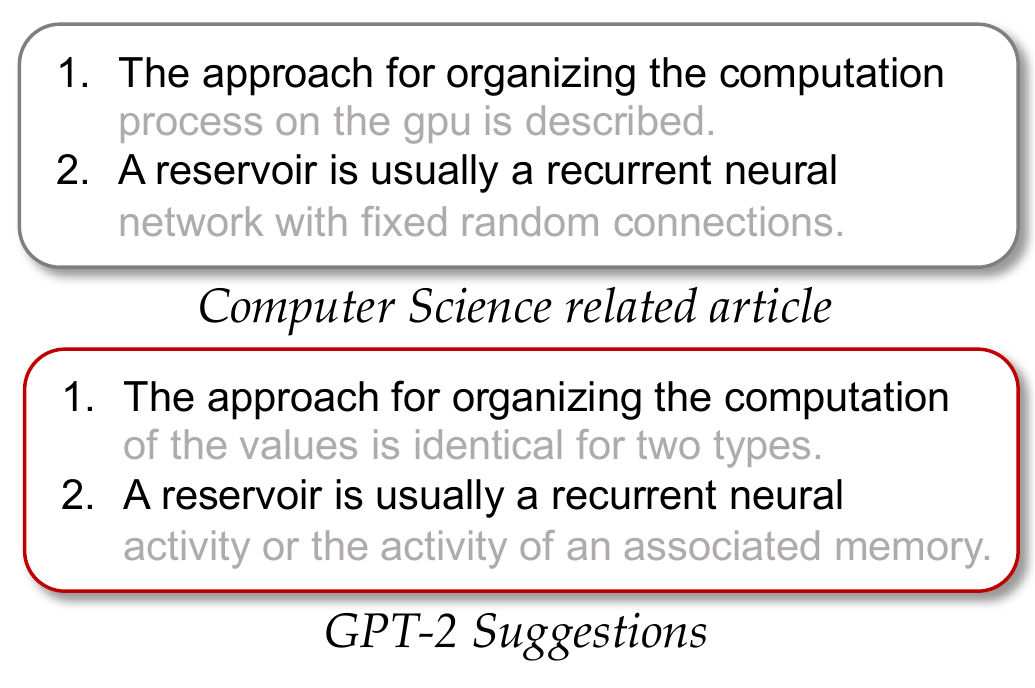}
	\caption{\textbf{Comparison of generated outputs.} GPT-2 can generate syntactically sound, and semantically general sentence from partial query. However, it still needs to be fine-tuned a lot to generate semantically expert domain (e.g. Computer Science) focused sentence.}
	\label{fig:example}
\end{figure}

A potential solution to address the challenges in text auto-completion is exploiting the Decoder-only Transformer model such as GPT-2~\cite{radford2019language}.
The model performs well on constructing syntactically sound sentences from a partial query.
However, GPT-2 requires a huge fine-tuning effort to construct sentences of expert domains.
~\figref{fig:example} shows an example of GPT-2 auto-completion suggestions for computer science domain sentences before fine-tuning. 
Recently, text-to-text transformers such as BART~\cite{lewis-etal-2020-bart} and T5~\cite{JMLR:v21:20-074} have demonstrated great potential in natural language generation (NLG) tasks by using masked-span infilling as a pre-training objective. 
However, similar to GPT-2, these models also require huge fine-tuning efforts to perform domain-specific text auto-completion.

This paper aims to address this research gap by proposing an intermediate training strategy~\cite{pruksachatkun-etal-2020-intermediate, calm2021}, which incrementally trains a pre-trained text-to-text transformer to provide better auto-completion suggestions and fastly adapt to the expert domain during fine-tuning. 
As shown in Figure~\ref{fig:overview}, the core of our intermediate training strategy is a simple self-supervised objective called \textit{Next Phrase Prediction (NPP)}, which has two major steps: Phrase Extraction (Section~\ref{sec:phrase}) and Generative Question Answering (Section~\ref{sec:qa}). 
The first stage extracts qualitative phrases by constituency parsing. 
By exploiting constituency parsing, the framework is able to utilize the complete phrase, not just a fraction of the sentence.
Next, the pre-trained language model is guided to choose the correct next phrase among other phrases of the same type (e.g., noun phrase, verb phrase, etc.) in the sentence.
For example, the sentence \textit{"She bought a top and bottom from that strange little shop."} has two noun phrases \textit{"a top and bottom"} and \textit{"that strange little shop"}. 
If the partial query is \textit{"She bought"}, the model is guided choose the proper complete noun phrase \textit{"a top and bottom"} for its next phrase.

To the best of our knowledge, this is the first work that proposed an intermediate training strategy for improving language models' performance on the text auto-completion task. 
Through extensive experiments, we demonstrated that our proposed approach could improve the text-to-text transformer's performance on auto-completion task and fastly adapt to expert domain of text auto-completion.


\section{Overview}
In this section, we first formalize the auto completion problem, and then introduce the workflow of our intermediate training strategy.

\subsection{Problem Statement}
Given partial query $\mathbf{p}=[s^{(1)}, s^{(2)}, \cdots, s^{(n)}]$, an auto completion returns $\mathbf{q}=[t^{(1)}, t^{(2)}, \cdots, t^{(n)}]$, where $\mathbf{q}$ is a syntactic and semantic extension of $\mathbf{p}$.
Specifically, every tokens of $\mathbf{p}$ is a prefix of $\mathbf{q}$, and every tokens of $\mathbf{q}$ is a suffix of $\mathbf{p}$; $[\mathbf{p};\mathbf{q}]$ is a full sentence.
We evaluate the auto completion model's performance on two attributes: (a) the soundness of $\mathbf{q}$, and (b) the semantic similarity of $\mathbf{q}$ with the ground truth.

\subsection{Workflow}
The workflow consists of two main steps, starting with a pre-trained T5 model: (i) applying the proposed self-supervised objective \textit{NPP} for intermediate-task training, and (ii) fine-tuning on target auto-completion task.

\begin{figure}[tb!]
\centering
	\includegraphics[width=0.9\linewidth]{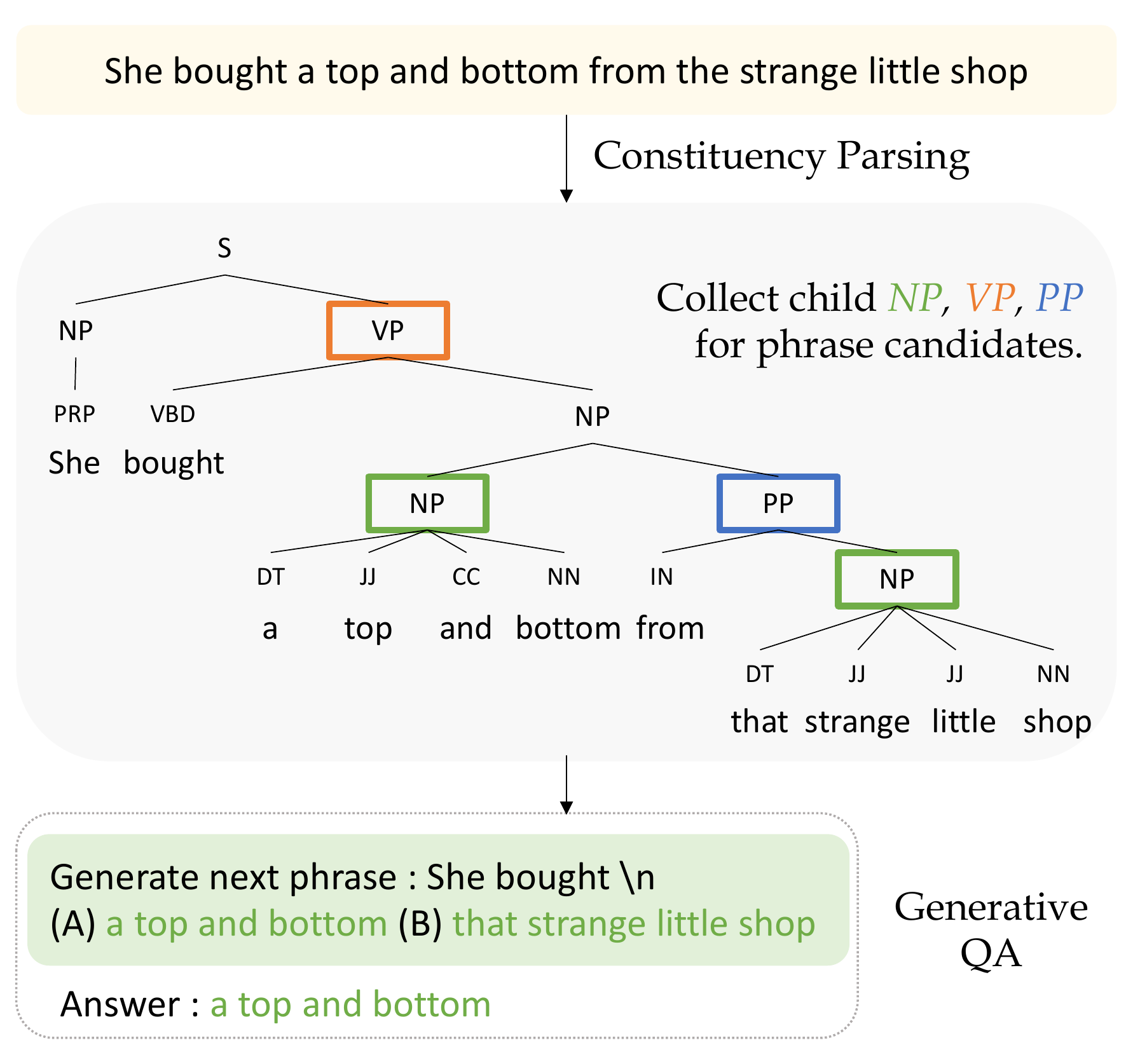}
	\caption{\textbf{Overview of next phrase prediction.} From the constituent tree, we retrieve the child phrases and group them according to their types (i.e., noun phrase (NP), verb phrase (VP), preposition phrase (PP).) Next, we randomly select a group that contains more than two phrases. Finally, We construct a generative QA style instance, where the phrases in the group are options to be selected as the correct next phrase for the input phrase.}
	\label{fig:overview}
\end{figure}
\section{Next Phrase Prediction}
The key idea of the \textit{next phrase prediction (NPP)} objective is to train a text-to-text transformer to complete the partial query with adequate phrases.
The underlying intuition of our proposed approach is as follows:
(1) Phrases tend to express meaning beyond simple word concatenation. For example, noun phrase such as \textit{"Recurrent Neural Network"} is constructed by three different words (\textit{"Recurrent"}, \textit{"Neural"}, \textit{"Network"}), where each word has its own meaning.
(2) Common phrases tend to be used on their own in the text. For instance, the prepositional phrase such as \textit{"in this paper"} frequently appears in academic writing domain.
Unlike existing language models that are trained to neglect such characteristics of phrases and predict the next word or span of the text, text auto-completion can be improved by performing phrase-level text completion as an intermediate training strategy in an effort to make the most of the phrase.
Specifically, \textit{NPP} involves two main steps: (i) Phrase Extraction, and (ii) Generative Question Answering (QA).

\subsection{Phrase Extraction}
\label{sec:phrase}
We first begin by extracting phrases using constituency parsing to retrieve qualitative phrases.
Given an input $\mathbf{x} = [x_{1},x_{2},\dots,x_{n}] $, we first conduct constituency parsing using AllenNLP~\cite{Gardner2017AllenNLP} and extract the \textit{Noun Phrase} (NP), \textit{Verb Phrase} (VP), and \textit{Prepositional Phrase} (PP). 
The extracted phrases are grouped into sets according to their types, denoted as $\mathcal{S}_{vp}, \mathcal{S}_{np}$, and $\mathcal{S}_{pp}$, respectively: $\mathcal{S}_{vp} = [vp_{1},vp_{2},\dots,vp_{q}], \mathcal{S}_{np} = [np_{1},np_{2}\dots,np_{q}]$, and $\mathcal{S}_{pp} = [pp_{1},pp_{2}\dots,pp_{q}]$.
For each phrase, we only keep the node that does not have a child node of the same phrase type.
For example, the sentence \textit{"She wants to eat pie."} has three VPs as follows:

\begin{minipage}[t]{0.5\textwidth}
    \vspace{0.2mm}
    \begin{flushleft}
        \small 
        (1) \textit{wants to eat pie (VP) $\rightarrow$ wants (VBZ) to eat pie (VP)} \\
        (2) \textit{to eat pie (VP) $\rightarrow$ to (TO) eat pie (VP)} \\
        (3) \textit{eat pie (VP)} \\
    \end{flushleft}
    \vspace{1mm}
\end{minipage}
To construct $\mathcal{S}_{vp}$ for this sentence, we only consider \textit{"eat pie"} as $vp_{i}$ to avoid word overlap between phrases.


\subsection{Generative QA}
\label{sec:qa}
After retrieving the phrases, we train the language model to predict the correct next phrase in a generative QA task setting~\cite{2020unifiedqa}. 
Specifically, from $\mathcal{S}_{vp}, \mathcal{S}_{np}$, and $\mathcal{S}_{pp}$, we randomly choose a set $\mathcal{S}$ that has more than two phrases.
To formulate the Generative QA task with the selected $\mathcal{S}$, here we present both the question and answer:
If the answer is a randomly chosen phrase $p$ from $\mathcal{S}$, then the question is composed of partial query $\mathbf{p}$ in which the chosen phrase $p$ is an extension of $\mathbf{p}$ and all phrases in $\mathcal{S}$ as answer choices.
The model is trained to output the correct phrase $p$, given partial query $\mathbf{p}$ and answer choices $\mathcal{S}$.
Figure~\ref{fig:overview} shows a real example of this format by choosing $\mathcal{S}_{np}$ as $\mathcal{S}$, \textit{"a top and bottom"} as $p$, and \textit{"She bought"} as $\mathbf{p}$.



\section{Experiments}

\subsection{Details for intermediate training}
We train a pre-trained T5-base model with \textit{NPP}.
We randomly sample 1M sentences from the English Wikipedia corpus\footnote{https://dumps.wikimedia.org/enwiki/latest/}, which is used for pre-training BERT and its variants, as the source data for \textit{NPP}.
The corpus has about 1.2B tokens, which is considerably less than the 34B token used in T5, and 10B tokens used in GPT2\footnote{Assuming the average token size is four characters.}.

\subsection{Target Dataset}
To show the effectiveness of our proposed method, we utilize two domains of text corpus to create the text auto-completion datasets:
\begin{itemize}
    \item \textbf{Email}: We utilize Enron email corpus~\footnote{https://www.cs.cmu.edu/\~./enron/} for general domain which is written in English collected from internal communication within a large business organization.
    \item \textbf{Academic writing}: We collect the abstracts of academic articles from  ArnetMiner~\cite{tang2008arnetminer}. The articles are written in English and mainly from the Computer Science domain, which are extracted from DBLP~\footnote{https://dblp.org/}, ACM~\footnote{https://www.acm.org/}, etc.
\end{itemize}
Table~\ref{tab:data} summerizes the statistics of the datasets used in our experiments. For data processing, we first extract the sentences from these text corpus.
For each sentence, we split into pairs $(\mathbf{p},\mathbf{q})$ by all word points. We consider $\mathbf{p}$ as partial phrase query to predict completion of the remaining phrase $\mathbf{q}$ in the sentence. Note that the $(\mathbf{p},\mathbf{q})$ formulation is used in fine-tuning the base models.

\begin{table}[!tb]
\begin{minipage}{0.5\textwidth}
\centering
\small

\resizebox{0.8\textwidth}{!}{
	\begin{tabular}{c|ccc}
		\toprule
		  \textbf{Dataset} & \textbf{Train} & \textbf{Dev} & \textbf{Test} \\
		\midrule
		  Emails & 156,998 & 13,474 & 15,030\\ 
		  Academics & 161,885 & 20,206 & 19,953\\ 
		\bottomrule
	\end{tabular}
	}
	\end{minipage}
    \vspace{0.1cm}
	\caption{\textbf{Statistics of datasets.}}
	\label{tab:data}
\end{table}

\begin{table*}[t!]
	\centering
		\resizebox{0.9\textwidth}{!}{
		\begin{tabular}{ccccccccc}
			\toprule
			\multirow{2}{*}{\textbf{Model / Metrics}}&
            \multicolumn{4}{c}{\textbf{Emails}} & \multicolumn{4}{c}{\textbf{Academic Writing}}\\
            \cmidrule(lr){2-5} \cmidrule(lr){6-9}
			  & BLEU-4 & METEOR & CIDEr & SPICE  & BLEU-4 & METEOR & CIDEr & SPICE  \\
			\midrule
            GPT-2~\cite{radford2019language} & 1.1 & 6.6 & 26.4 & 3.3 & 0.6 & 6.0 & 23.6 & 2.6 \\
            T5~\cite{JMLR:v21:20-074} & 2.8 & 6.8 & 39.8 & 4.2 & 2.2 & 7.5 & 50.3 & 3.9 \\
            \midrule
            NSP+T5 & 3.0 & 6.9 & 41.1 & 4.4 & 2.3 & 7.5 & 51.1 & 4.0 \\
            NPP+T5 (Ours) & \bf 3.2 & \bf 7.1 & \bf 43.0 & \bf 4.5 & \bf 2.5 & \bf 7.8 & \bf 53.5 & \bf 4.2\\
			\bottomrule
		\end{tabular}
		
		}
		\vspace{0.1cm}
		\caption{{\textbf{Experimental Results}}. The first group of models are baselines which are not intermediately trained. Last group of models are intermediately-trained with different objectives. Best models are bold within each metric.}
	\label{tab:experiments}
\end{table*}

\begin{table*}[t!]
	\centering
	
		\resizebox{\textwidth}{!}{
		\begin{tabular}{l|l|l|l}
			\toprule
			\textbf{Partial Query} & \textbf{Original} & \textbf{T5} & \textbf{NPP+T5} \\
			\midrule
            Building large OCR databases is a time 
            & \underline{consuming} and tedious work .
            & challenging .
            & \underline{consuming} task . \\
            
            vpi is part of the ieee programming & \underline{language} interface standard . & system . & \underline{language} .\\
            
            a connection between the kalman & \underline{filter} is developed . & et al . & \underline{filter} is established . \\
            appendix provides a complete listing & of code for the systems . & of the apl libraries . & of the tools and techniques used in this paper . \\
            automatic target & recognition is an important task . & selection is based on a set of criteria . &  detection is a key feature of this approach . \\
			\bottomrule
		\end{tabular}
		}
		\vspace{0.1cm}
	\caption{\textbf{Generated Examples of Academic Writing}. For the same partial queries from academic writing dataset, we compare the generated completions between T5 and NPP+T5. Underlines are overlap words between the original completion and generated completions.}
	\label{tab:generations}
	\vspace{-0.5cm}
\end{table*}
\subsection{Base models}
We compare our proposed approach with other pre-trained language generation models.
We fine-tuned the following models on our training data in a sequence-to-sequence format:
(1) \textbf{GPT-2}~\cite{radford2019language} is the pre-trained GPT-2 large model, which has 774M parameters.
For fine-tuning, we condition the model on the format $\mathbf{p}=\mathbf{q}$.
For inference, we sample from the fine-tuned GPT-2 model after a prompt of the partial query $\mathbf{p}$ with beam search, and cleaning the samples by postprocessing. Then, we use the first sample as the output sentence.
(2) \textbf{T5}~\cite{JMLR:v21:20-074} is the pre-trained T5-base model, which has 220M parameters.
For fine-tuning, we prepend the prefix: \texttt{"generate next phrase:"} to partial query $\mathbf{p}$ and feed into the model to generate completion $\mathbf{q}$.
(3) \textbf{NSP+T5} is intermediately-trained based on T5-Base using Next Sentence Prediction (NSP), which is used in BERT~\cite{devlin-etal-2019-bert} pre-training.
(4) \textbf{NPP+T5} is intermediately-trained based on T5-base using Next Phrase Prediction (NPP), which is our proposed objective.

\subsection{Evaluation Metrics}
To evaluate the syntactic and semantic soundness of generated sentences, we exploit several widely used automatic metrics to assess the performance, such as BLEU~\cite{papineni-etal-2002-bleu}, METEOR~\cite{banerjee-lavie-2005-meteor}, CIDEr~\cite{vedantam2015cider}, and SPICE~\cite{spice2016}.
These metrics evaluate whether the model is able to generate semantically expert domain focused sentence by measuring surface similarities and associations between system generations and original text.

\subsection{Experimental Results}
Table~\ref{tab:experiments} shows the experimental results of text auto-completion on the email and academic writing datasets. We observed that the model intermediately trained with our objective outperforms the base models on both datasets.

Specifically, our approach, \textbf{NPP+T5}, outperforms \textbf{NSP+T5} by a margin from 0.2 to 0.3 BLEU/METEOR/SPICE score, suggesting that predicting the next phrase is more effective than predicting next sentence in text auto-completion task.
Moreover, we also observe that \textbf{NPP+T5} outperforms \textbf{GPT-2} even though the number of parameters in \textbf{NPP+T5} is less than half of \textbf{GPT-2}.
The experimental results demonstrated the flexibility of our proposed approach, which can serve as "plug-and-play" for any text-to-text transformer models and enhance their performance in the text auto-completion task.

~\tabref{tab:generations} shows the comparison of generated suggestions for the same partial query between \textbf{T5} and \textbf{NPP+T5}. We can observe that the completions by \textbf{NPP+T5} are generally more acceptable in terms of semantic similarity between generated completions and original text.

\section{Conclusion}
In this paper, we propose a novel intermediate training strategy that encourages the model to complete the partial query with enriched phrases and eventually improving the performance of the text auto-completion system.
Our proposed approach enhances state-of-the-art language model's performance by intermediately training it with our next phrase prediction self-supervised objective. 
Preliminary experiments have shown that our approach is able to outperform the baselines in auto-completion for email and academic-writing domains with only around 1.2B tokens of training.
For future work, we aim to experiment our proposed approach on text auto-completion in more writing domains and develop a demonstration system to better showcase our approach in text auto-completion.
\bibliographystyle{acl_natbib}
\bibliography{emnlp2021}

\end{document}